\documentclass[runningheads, article]{llncs}
\usepackage{amsfonts}
\usepackage{amscd}
\usepackage{amssymb}
\usepackage{amsmath}
\usepackage{bm}
\usepackage{mathrsfs}
\usepackage{mathtools}
\usepackage{here}
\usepackage{examplep}
\usepackage{mdframed}
\usepackage{graphicx}
\usepackage[T1]{fontenc}
\usepackage{color}

\usepackage{booktabs}
\usepackage{comment}
\usepackage{tabularx}
\usepackage[hidelinks]{hyperref}
\usepackage{subcaption}

\begin{document}

\title{Interpretable Clustering \\ via Optimal Multi-way Decision Trees}

\titlerunning{Interpretable Clustering via Optimal Multi-way Decision Trees}

%\begin{comment}
\author{Hayato Suzuki\inst{1}
% , \orcidID{0000-1111-2222-3333} 
\and
Shunnosuke Ikeda\inst{2}
% \orcidID{0009-0004-4283-0819} 
\and 
Naoki Nishimura\inst{3}
\and 
Yuichi Takano\inst{4}
% \orcidID{0000-0002-8919-1282}
}
\authorrunning{H. Suzuki et al.}

\institute{Graduate School of Science and Technology, University of Tsukuba, \\ 1--1--1 Tennodai, Tsukuba-shi, Ibaraki 305--8573, Japan \\
\email{s2420503@u.tsukuba.ac.jp}\\
\and
Institute of Systems and Information Engineering, University of Tsukuba, \\ 1--1--1 Tennodai, Tsukuba-shi, Ibaraki 305--8573, Japan \\
\email{ikeda@cs.tsukuba.ac.jp}\\
\and
Data Management \& Planning Office, Recruit Co., Ltd., \\ 1--9--2 Marunouchi, Chiyoda-ku, Tokyo 100--6640, Japan \\
\email{nishimura@r.recruit.co.jp}\\
\and
Institute of Systems and Information Engineering, University of Tsukuba, \\ 1--1--1 Tennodai, Tsukuba-shi, 305--8573, Ibaraki, Japan \\
\email{ytakano@sk.tsukuba.ac.jp}}
%
%\end{comment}

\maketitle 
\begin{abstract}
%150words～200words
Clustering is a fundamental unsupervised learning technique for uncovering data structures to facilitate knowledge discovery and decision-making. 
While clustering accuracy is crucial, interpretability significantly impacts the practical value of clustering results, particularly in high-risk decision-making contexts. 
Although decision-tree-based clustering methods offer high interpretability through explicit splitting rules, existing approaches often rely on local greedy search or require expensive computational costs limited to binary splits, resulting in deeper, less interpretable trees. 
To overcome these limitations, we establish a high-performance computational framework named Interpretable Clustering via Optimal Multi-way Trees (ICOMT). 
We make three primary contributions. First, we propose a new discretization method for numerical features using one-dimensional $K$-means clustering to capture data distributions. 
Second, we formulate a binary linear optimization (BLO) problem to guarantee tree optimality. 
Third, extensive validation on four public datasets demonstrates that our ICOMT method outperforms existing baselines, achieving superior clustering accuracy while maintaining shallow, concise tree structures.

\end{abstract}

\section{Introduction}\label{sec:intro}

\subsection{Background}\label{subsec:background}
Clustering serves as a fundamental unsupervised learning technique that organizes data points into groups with similar characteristics.
Its primary purpose involves uncovering underlying data structures to facilitate knowledge discovery and decision-making. 
The output of clustering is typically represented as a partition (i.e., clusters) of the dataset.
Clustering accuracy is a measure of how accurately these clusters represent the underlying structure of the dataset, and is therefore crucial because it directly impacts the quality of subsequent data analysis.
Additionally, when clustering is performed as a preprocessing step in predictive analytics, clustering accuracy also affects the performance of the resultant predictive model.

However, accuracy is not the only factor determining the practicality of clustering results.
The interpretability of clustering results also significantly impacts their value, especially in high-risk decision-making situations~\cite{rudin2019stop}.
This concept refers to the ability to understand the characteristics of each cluster and explain the resultant structure~\cite{Forgy1965}.
For example, stratifying patients with similar characteristics in medical settings can facilitate clinical decision-making by segmenting diseases, predicting treatment responses, personalizing patient care, and streamlining clinical trials~\cite{ahlqvist2018novel,windgassen2018importance}.

Despite the recognized importance of clustering interpretability, research that directly addresses this challenge remains limited~\cite{hu2026}. 
Therefore, the development of clustering methods that can simultaneously achieve high accuracy and interpretability has emerged as a vital research direction.
The need for machine learning methods that combine both accuracy and interpretability has also been emphasized by regulatory bodies such as the European Union~\cite{Goodman2016}.

\subsection{Related Work}\label{subsec:relatedwork}

A wide variety of clustering methods have been proposed~\cite{Jain1999}, including $K$-means clustering, hierarchical clustering, model-based clustering, and density-based clustering. 
While representative points are widely used for interpreting clusters~\cite{Diday1976}, they may fail to fully capture the essential characteristics of clusters exhibiting elongated shapes or anisotropic distributions~\cite{Jain1999}.
Another common strategy is to visualize data points by projecting them into a two-dimensional space using principal component analysis~\cite{Jolliffe2011}; however, this method can obscure the correspondence between clusters and the original features.
In contrast, decision-tree-based clustering methods are highly interpretable because they represent cluster characteristics explicitly through splitting rules.

A simple way for applying decision trees to clustering is a two-step scheme: clustering is performed first, and then a decision tree is trained to classify the obtained cluster labels~\cite{Hancock2003,Jain1999,moshkovitz2020}. 
However, since cluster formation and decision tree construction are performed within different frameworks, the trained decision tree may fail to faithfully reflect the actual cluster structure.
To overcome this limitation, various algorithms have been proposed that incorporate interpretability directly into the cluster formation step; for a systematic review of these algorithms, see Hu et al.~\cite{hu2026}.

Predictive Clustering Trees (PCT)~\cite{Blockeel1998} are a general-purpose framework for decision tree construction applicable to both supervised and unsupervised learning. 
Clustering using Unsupervised Binary Trees (CUBT)~\cite{Fraiman2013} is a clustering tree construction method consisting of a series of recursive binary splits, aggregation of adjacent nodes, and joining of similar clusters.
However, many of these tree construction methods are based on local greedy search and therefore cannot guarantee the optimality of the resultant decision tree. 

Building on these lines of work, Bertsimas et al.~\cite{bertsimas2021} proposed Interpretable Clustering via Optimal Trees (ICOT) by extending optimal classification trees~\cite{bertsimas2017optimal} to unsupervised learning. 
This method uses mixed-binary optimization to perform clustering and tree construction simultaneously.
However, ICOT is computationally expensive because it requires solving a mixed-binary nonlinear optimization problem. 
Furthermore, its limitation to binary decision trees tends to generate deeper tree structures, leading to reduced interpretability.

On the other hand, Subramanian et al.~\cite{subramanian2023} proposed Optimal Multi-way Trees (OMT) for supervised learning.
This method uses binary linear optimization to construct an optimal multi-way decision tree by selecting a subset of paths on a feature graph that represents the feature splitting conditions.
Such multi-way decision trees have been reported to generate more concise splitting rules than binary decision trees and thus maintain high interpretability even with shallow tree structures.
However, since OMT is designed for supervised learning (e.g., regression and classification), it is not applicable to unsupervised learning tasks such as clustering.

\subsection{Our Contribution}\label{subsec:contribution}
The goal of this research is to establish a high-performance computational framework for interpretable clustering, named Interpretable Clustering via Optimal Multi-way Trees (ICOMT), by extending optimal multi-way decision trees~\cite{subramanian2023} to clustering tasks.
The main contributions of this paper can be outlined as follows:
\begin{itemize}
  \item \textbf{New discretization method for numerical features}:\\
  In the previous study~\cite{subramanian2023} on OMT, numerical features were discretized for multi-way tree construction using quantile-based cumulative binning, but this method has the drawback of not adequately reflecting the shape of the data distribution.
  To overcome this challenge, we propose the use of one-dimensional $K$-means clustering for discretizing numerical features. 
\vspace{1mm}
  \item \textbf{BLO formulation for optimal multi-way clustering trees}:\\
  By extending the OMT framework~\cite{subramanian2023}, designed for supervised learning, to unsupervised learning, we formulate a binary linear optimization (BLO) problem for constructing multi-way clustering trees. 
  As a result, we can establish a comprehensive procedure for constructing multi-way decision trees for interpretable clustering.
\vspace{1mm}
  \item \textbf{Performance validation on publicly available datasets}:\\
  Through experiments using four publicly available datasets, we confirm that our ICOMT method outperforms existing baseline methods (ICOT~\cite{bertsimas2021}, PCT~\cite{Blockeel1998}, and the two-step (cluster formation and tree construction) scheme~\cite{moshkovitz2020}) in terms of both clustering accuracy and interpretability.
\end{itemize}

\section{Methods}\label{sec:method}
In this section, we present our ICOMT framework for computing optimal multi-way decision trees for interpretable clustering. 
Throughout this paper, we denote the set of consecutive positive integers as $[n] \coloneqq \{1, 2, \ldots, n\}$.

\begin{figure}[!t]
    \centering
    % --- 左側の図 ---
    \begin{subfigure}{0.49\linewidth}
        \centering
        \includegraphics[width=\linewidth]{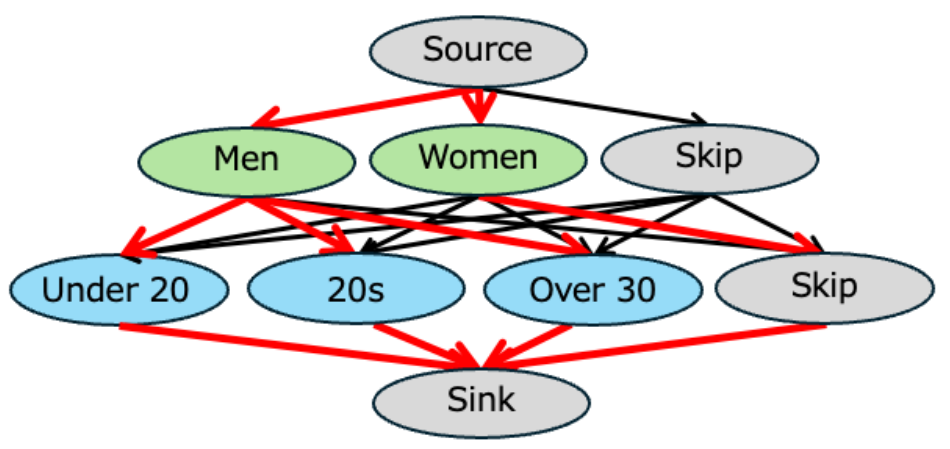}
        \caption{Feature graph}
        \label{fig:example_graph}
    \end{subfigure}
    \hfill % 図の間に適切なスペースを挿入
    % --- 右側の図 ---
    \begin{subfigure}{0.49\linewidth}
        \centering
        \includegraphics[width=\linewidth]{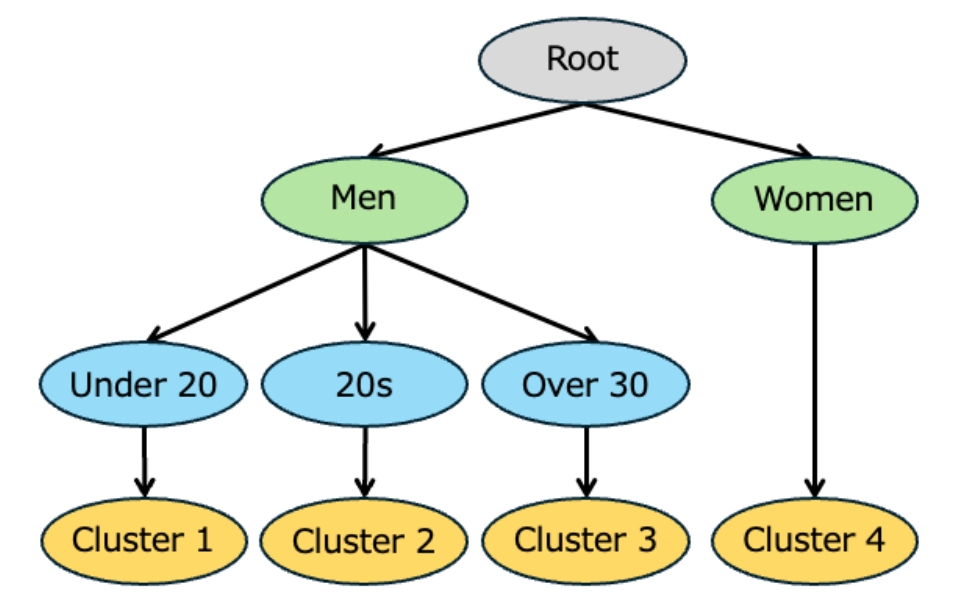}
        \caption{Multi-way decision tree}
        \label{fig:example_tree}
    \end{subfigure}

    \caption{Feature graph for constructing multi-way decision trees}
    \label{fig:example}
\end{figure}

\subsection{Overview}\label{subsec:overview}

We will now provide an overview of our interpretable clustering method with multi-way decision trees.
This method employs a directed acyclic graph called a \emph{feature graph}~\cite{subramanian2023}, where each feature constitutes a level in the graph, and each level consists of multiple nodes corresponding to different feature values.
Nodes in a given level are fully connected to nodes in adjacent levels.

Fig.~\ref{fig:example_graph} illustrates an example feature graph with two features (\texttt{Gender} and \texttt{Age}).
For categorical features like \texttt{Gender}, each category directly corresponds to a node.
On the other hand, in the case of numerical features like \texttt{Age}, a finite number of intervals are created beforehand and treated as nodes.
Each directed path from the \texttt{Source} node to the \texttt{Sink} node corresponds to a sequence of feature splitting conditions (e.g., $X_1 = \mathtt{Men}$ and $X_2 = \mathtt{Under~20}$).
Here, the \texttt{Skip} nodes indicate that the corresponding feature is not used in the splitting conditions.

A multi-way decision tree can be constructed by selecting a subset of paths from the \texttt{Source} node to the \texttt{Sink} node on a feature graph.
For example, selecting the paths composed of red edges in Fig.~\ref{fig:example_graph} yields the multi-way decision tree in Fig.~\ref{fig:example_tree}, where each path (or leaf node) corresponds to a cluster, defined by a set of data points that satisfy feature conditions along the path from the \texttt{Root} node. 
In other words, each path in a multi-way decision tree corresponds to an interpretable clustering rule, making it possible to represent the clustering results as a clear logical structure.

\subsection{Discretization of Numerical Features}\label{subsec:discretization}

Let us suppose that there are $p$ features to use for clustering. 
In the previous study~\cite{subramanian2023}, numerical features were discretized using the \emph{cumulative binning method} based on quantile discretization. 
Specifically, for each numerical feature $X_f~(f \in [p])$, two thresholds $Q_{f1}$ and $Q_{f2}$ are determined ($Q_{f1} < Q_{f2}$) such that the number of data points in each of the three resultant intervals is approximately equal.
Next, by adding combinations of adjacent intervals, the following five (partially overlapping) intervals are defined: 
\begin{equation}\label{eq:qumBin}
    (-\infty, Q_{f1}), \quad
    [Q_{f1}, Q_{f2}), \quad
    [Q_{f2}, \infty), \quad 
    (-\infty, Q_{f2}), \quad
    [Q_{f1}, \infty).
\end{equation}

While this cumulative binning method is effective in keeping the number of data points within each interval nearly equal, it has the drawback of not considering the shape of the data distribution.
Therefore, if the data distribution is skewed or multimodal, the resultant intervals may fail to accurately reflect the underlying data structure. 

To overcome this challenge, we introduce a \emph{1D-$K$-means binning method} using one-dimensional $K$-means clustering. 
Specifically, each numerical feature $X_f~(f \in [p])$ is divided into $K_f$ clusters using $K$-means clustering, where the centroid of each cluster is represented by $\bar{x}_{fk}$ for $k \in [K_f]$.
Here, the number of clusters $K_f \in \{2, 3, \ldots, 6\}$ is chosen to minimize the Bayesian information criterion (BIC).
Next, the interval threshold is set to the midpoint of the centroids of adjacent clusters:
\[
Q_{fk} \coloneqq \frac{\bar{x}_{fk} + \bar{x}_{f,k+1}}{2}, \quad \forall k \in [K_f - 1].
\]
Then, similarly to the cumulative binning method (Eq.~\eqref{eq:qumBin}), we define $2 K_f - 1$ (partially overlapping) intervals by combining adjacent intervals.

\subsection{BLO Formulation}\label{subsec:BLOformulation}

We will now present our mathematical optimization formulation for constructing multi-way decision trees with guaranteed optimality for interpretable clustering. 
Recall that each path on the feature graph corresponds to a cluster, and the objective is to select an optimal subset of paths for constructing multi-way clustering trees (Fig.~\ref{fig:example}).
Table~\ref{tab:notation_icomt} lists the notation used in our formulation.

\setlength{\tabcolsep}{3pt}
\begin{table}[t]
\centering
\caption{Notation used in our formulation}
\label{tab:notation_icomt}
\begin{tabular}{ll}
\toprule
Symbol & Description \\
\midrule
%\multicolumn{2}{l}{\textbf{Index sets:}} \\
%    \quad $I \coloneqq \{1,2,\dots,N\}$ & Set of data points \\
%    \quad $J$ & Set of paths (or candidate clusters) \\
%\addlinespace[0.5em]
\multicolumn{2}{l}{\textbf{Decision variables:}} \\
    \quad $z_j \in \{0,1\}$ & $z_j = 1$ iff path $j \in [m]$ is selected \\
\addlinespace[0.5em]
\multicolumn{2}{l}{\textbf{Constants:}} \\
    \quad $a_{ij} \in \{0,1\}$ & $a_{ij} = 1$ iff data point $i \in [n]$ corresponds to path $j \in [m]$ \\
    \quad $c_j$ & Cost of path $j \in [m]$ defined by Eq.~\eqref{eq:xi} \\
    \quad $m$ & Number of paths (or candidate clusters) \\
    \quad $n$ & Total number of data points \\
    \quad $n_j$ & Number of data points corresponding to path $j \in [m]$ \\
    \addlinespace[0.5em]
\multicolumn{2}{l}{\textbf{Parameters:}} \\
    \quad $\kappa$ & Upper limit of the number of clusters \\
    \quad $\rho \in [0,1]$ & Lower limit of data point coverage \\
\bottomrule
\end{tabular}
\end{table}

For each data point $i \in [n]$, let $\bm{x}_i \in \mathbb{R}^p$ be the $p$-dimensional feature vector.
Let $D_j \coloneqq \{i \in [n] \mid a_{ij} = 1\}$ be the index set of data points corresponding to path $j \in [m]$.
The cost of each path $j \in [m]$ is then defined by the associated within-cluster sum of squares as 
\begin{equation}
c_j \coloneqq \sum_{i \in D_j} \|\bm{x}_i - \bm{\mu}_j\|^2, \quad \forall j \in [m]
\label{eq:xi}
\end{equation}
where $\bm{\mu}_j \coloneqq (\sum_{i \in D_j} \bm{x}_i)/|D_j| \in \mathbb{R}^p$ is the centroid of the data points corresponding to path $j \in [m]$.
Notably, this cost can be calculated after enumerating the candidate paths on the feature graph, and can therefore be treated as a constant in the following optimization problem.

Let $\bm{z} \coloneqq (z_j)_{j \in [m]} \in \{0,1\}^m$ be the binary decision variable vector for path selection, where $z_j = 1$ holds if and only if path $j \in [m]$ is selected. 
Our method for constructing multi-way clustering trees is then formulated as the following binary linear optimization (BLO) problem: 
\begin{align}
\min_{\bm{z}} \quad & \sum_{j \in [m]} c_j z_j \label{eq:objective} \\
\text{s.~t.} \quad
& \sum_{j \in [m]} a_{ij} z_j \le 1, \quad \forall i \in [n], \label{eq:assign}\\
& \sum_{j \in [m]} z_j \le \kappa, \label{eq:leaf}\\
& \sum_{j \in [m]} n_j z_j \ge \rho n, \label{eq:coverage}\\
& z_j \in \{0,1\}, \quad \forall j \in [m], \label{eq:binary}
\end{align}
where
\begin{itemize}
  \item Eq.~\eqref{eq:objective}: 
  Minimize the total within-cluster sum of squares (cf.~Eq.~\eqref{eq:xi}) associated with the selected paths (or clusters) to improve cluster cohesion;
  \item Eq.~\eqref{eq:assign}:   
  Ensure that each data point is covered by at most one path (or cluster); 
  \item Eq.~\eqref{eq:leaf}:   
  Ensure that the number of selected paths (or clusters) is at most $\kappa$; 
  \item Eq.~\eqref{eq:coverage}:   
  Ensure that at least $\rho n$ data points are covered by the selected paths (or clusters); 
  \item Eq.~\eqref{eq:binary}:   
  Ensure that $z_j$ is a binary decision variable for path selection for each $j \in [m]$.
\end{itemize}

\subsection{Algorithm}\label{sec:algorithm}

We are now in a position to describe the overall algorithm of our ICOMT framework. 

\begin{enumerate}
  \item \textbf{Discretization of numerical features}: 
  Numerical features are discretized using the previous cumulative binning method~\cite{subramanian2023} or our 1D-$K$-means binning method (Section~\ref{subsec:discretization}).
  \item \textbf{Feature graph construction}: 
  The feature graph is constructed as shown in Fig.~\ref{fig:example_graph}, where each feature constitutes a level in the graph, and each level consists of multiple nodes corresponding to the (discretized) feature values.
  \item \textbf{Path enumeration on the feature graph}:  
  All paths from the \texttt{Source} node to the \texttt{Sink} node are enumerated using depth-first search on the feature graph, where each path corresponds to a candidate cluster. 
  \item \textbf{Elimination of redundant paths}: Redundant paths covering the same set of data points are eliminated to improve the computational efficiency of the subsequent optimization;
  \item \textbf{Path evaluation}: The within-cluster sum of squares (Eq.~\eqref{eq:xi}) is calculated to define the cost of each path. 
  \item \textbf{Optimal path selection}: 
  The BLO problem~\eqref{eq:objective}--\eqref{eq:binary} is solved to find an optimal subset of paths that minimize the total within-cluster sum of squares.  
  \item \textbf{Multi-way tree construction}:  
  A multi-way decision tree is constructed from the selected subset of paths as shown in Fig.~\ref{fig:example_tree}, and cluster labels are assigned to the data points according to the obtained decision tree. 
\end{enumerate}

\section{Experiments}\label{sec:experiments}
In this section, we evaluate the effectiveness of our ICOMT method for interpretable clustering through numerical experiments using publicly available datasets.
All experiments were performed on a MacBook Pro equipped with an Apple M3 Pro (11-core CPU) and 18 GB of unified memory.

\subsection{Datasets}\label{subsec:datasets}

We used the following four publicly available datasets, where $n$ is the number of data points, $p$ is the number of features, and $K^{\star}$ is the number of ground-truth cluster labels: 

\begin{itemize}
\item \textbf{Seeds}~\cite{charytanowicz2010seeds} ($n=210,~p=7,~K^{\star}=3$): This dataset consists of 210 samples of wheat kernels, each described by seven geometric attributes extracted from soft X-ray images. 
Three different varieties of wheat were used as cluster labels.
\vspace{1mm}
\item \textbf{Customer}~\cite{ksoy2024customer} ($n=238,~p=4,~K^{\star}=7$): 
This dataset contains synthetic purchasing behavior data for 238 customers, each with four attributes. 
Customer loyalty scores were rounded to integers between 3 and 9 for cluster labels.
\vspace{1mm}
\item \textbf{Estate}~\cite{yeh2018realestate} ($n=414,~p=6,~K^{\star}=3$): This dataset contains 414 real estate records from New Taipei City, Taiwan, each described by six attributes. 
House price of unit area was transformed into three cluster labels using one-dimensional $K$-means clustering, with cluster boundaries of 32 and 49.
\vspace{1mm}
\item \textbf{Statlog}~\cite{siebert1987vehicle} ($n=845,~p=18,~K^{\star}=4$): This dataset consists of 845 samples described by 18 shape features (Table~\ref{tab:statlog_features}) extracted from 2D vehicle silhouettes. 
Four types of vehicle were used as cluster labels.
\end{itemize}
Note here that these datasets contain target variables for supervised learning, and the ground-truth cluster labels were defined using these target variables to evaluate clustering accuracy.
Prior to the analysis, min-max normalization was applied to rescale all features to the interval $[0,1]$. 

\begin{table}[t]
\centering
\caption{Features in the Statlog dataset}
\label{tab:statlog_features}
\begin{tabular}{llll}
\toprule
ID  & Description \\
\midrule
%Y   & Vehicle type (target label) \\
X1  & Degree of compactness of the silhouette \\
X2  & Degree of circularity \\
X3  & Circularity measure based on distances \\
X4  & Ratio of radii (shape dispersion) \\
X5  & Aspect ratio along the principal axis \\
X6  & Aspect ratio based on maximum length \\
X7  & Ratio capturing the spread/scatter of the shape \\
X8  & Degree of elongation \\
X9  & Rectangularity along the principal axis \\
X10 & Rectangularity along the maximum length \\
X11 & Scaled variance along the major axis \\
X12 & Scaled variance along the minor axis \\
X13 & Scaled radius of gyration (shape spread) \\
X14 & Skewness about the major axis \\
X15 & Skewness about the minor axis \\
X16 & Kurtosis about the minor axis \\
X17 & Kurtosis about the major axis \\
X18 & Ratio related to hollowness/concavity \\
\bottomrule
\end{tabular}
\end{table}

\subsection{Methods under Comparison}\label{subsec:methods}

We compared the performance of the following five methods for clustering tree construction:
\begin{itemize}
    \item \textbf{ICOMT-K}: Our BLO model~\eqref{eq:objective}--\eqref{eq:binary} for interpretable clustering via optimal multi-way trees using the 1D-$K$-means binning method (Section~\ref{subsec:discretization}); 
    \vspace{1mm}
    \item \textbf{ICOMT-C}:   
    Our BLO model~\eqref{eq:objective}--\eqref{eq:binary} for interpretable clustering via optimal multi-way trees using the cumulative binning method~\cite{subramanian2023}; 
    \vspace{1mm}
    \item \textbf{ICOT}:   
    Mixed-binary nonlinear optimization model~\cite{bertsimas2021} for interpretable clustering via optimal (binary) trees; 
    \vspace{1mm}
    \item \textbf{PCT}:   
    Predictive clustering tree method~\cite{Blockeel1998}, which treats the feature vector as the target variable and trains regression trees to minimize the within-cluster variance;
    \vspace{1mm}
    \item \textbf{2-Step}:   
    Two-step method~\cite{moshkovitz2020}, which first generates cluster labels using $K$-means clustering and then trains classification and regression trees (CART)~\cite{Breiman1984} using those as target labels.
\end{itemize}

For the ICOMT methods, the BLO problem~\eqref{eq:objective}--\eqref{eq:binary} was solved using Gurobi Optimizer\footnote{\url{https://www.gurobi.com}} 12.0.2, with the lower limit of data point coverage set to $\rho=1.0$. 
For the ICOT method, the coordinate descent heuristic~\cite{bertsimas2021} was implemented in Python, where the initial solution was obtained using the 2-Step method. 
The PCT and 2-Step methods were implemented using KMeans and DecisionTreeRegressor from the Scikit-learn library in Python. 
For all the methods, the maximum tree depth was set to three. 

\begin{table}[!p]
\centering
\caption{Results on clustering performance}
\label{tab:ari}

    \begin{subtable}{\linewidth}
    \centering
    \caption{Seeds dataset}
    \label{tab:ari1}
    \begin{tabular}{lrrrrrrrr}
    \toprule
    & \multicolumn{4}{c}{Tuned by silhouette coef.} & \multicolumn{4}{c}{Tuned by Dunn index} \\
    \cmidrule(lr){2-5} \cmidrule(lr){6-9}
    Method & ARI & Dep & \#Cls & Time (s) & ARI & Dep & \#Cls & Time (s) \\
    \midrule
    ICOMT-K       & \textbf{0.614} & 1 & 3  & 0.241 & \textbf{0.614} & 1 & 3  & 0.241 \\
    ICOMT-C       & 0.455          & 1 & 2  & 0.091 & 0.366          & 3 & 10 & 0.112 \\
    ICOT          & 0.486          & 1 & 2  & 0.098 & 0.571          & 3 & 5  & 0.227 \\
    PCT           & 0.486          & 1 & 2  & 0.001 & 0.486          & 1 & 2  & 0.001 \\
    2-Step        & 0.497          & 3 & 4  & 0.093 & 0.523          & 3 & 5  & 0.172 \\
    \bottomrule
    \end{tabular}
    \end{subtable}

    \begin{subtable}{\linewidth}
    \centering
    \caption{Customer dataset}
    \label{tab:ari2}
    \begin{tabular}{lrrrrrrrr}
    \toprule
    & \multicolumn{4}{c}{Tuned by silhouette coef.} & \multicolumn{4}{c}{Tuned by Dunn index} \\
    \cmidrule(lr){2-5} \cmidrule(lr){6-9}
    Method & ARI & Dep & \#Cls & Time (s) & ARI & Dep & \#Cls & Time (s) \\
    \midrule
    ICOMT-K       & \textbf{0.595} & 3 & 8 & 0.014 & \textbf{0.595} & 3 & 8 & 0.014 \\
    ICOMT-C       & 0.235          & 1 & 2 & 0.008 & 0.452          & 1 & 3 & 0.007 \\
    ICOT          & 0.550          & 3 & 5 & 0.068 & 0.550          & 3 & 5 & 0.070 \\
    PCT           & 0.551          & 3 & 5 & $< 0.001$ & 0.551          & 3 & 8 & $<0.001$ \\
    2-Step        & 0.502          & 3 & 6 & 0.042 & 0.502          & 3 & 6 & 0.036 \\
    \bottomrule
    \end{tabular}
    \end{subtable}

    \begin{subtable}{\linewidth}
    \centering
    \caption{Estate dataset}
    \label{tab:ari3}
    \begin{tabular}{lrrrrrrrr}
    \toprule
    & \multicolumn{4}{c}{Tuned by silhouette coef.} & \multicolumn{4}{c}{Tuned by Dunn index} \\
    \cmidrule(lr){2-5} \cmidrule(lr){6-9}
    Method & ARI & Dep & \#Cls & Time (s) & ARI & Dep & \#Cls & Time (s) \\
    \midrule
    ICOMT-K       & 0.195          & 1 & 4 & 0.585 & 0.142          & 2 & 8 & 0.540  \\
    ICOMT-C       & \textbf{0.251} & 1 & 2 & 0.146 & \textbf{0.251} & 1 & 2 & 0.146 \\
    ICOT          & 0.096          & 3 & 8 & 0.499 & 0.096          & 3 & 8 & 0.258 \\
    PCT           & 0.094          & 3 & 8 & 0.001 & 0.094          & 3 & 8 & 0.001 \\
    2-Step        & 0.096          & 3 & 8 & 0.478 & 0.096          & 3 & 8 & 0.231 \\
    \bottomrule
    \end{tabular}
    \end{subtable}

    \begin{subtable}{\linewidth}
    \centering
    \caption{Statlog dataset}
    \label{tab:ari4}
    \begin{tabular}{lrrrrrrrr}
    \toprule
    & \multicolumn{4}{c}{Tuned by silhouette coef.} & \multicolumn{4}{c}{Tuned by Dunn index} \\
    \cmidrule(lr){2-5} \cmidrule(lr){6-9}
    Method & ARI & Dep & \#Cls & Time (s) & ARI & Dep & \#Cls & Time (s) \\
    \midrule
    ICOMT-K       & \textbf{0.107} & 1 & 3  & 24.686 & 0.114          & 2 & 4  & 86.844 \\
    ICOMT-C       & 0.083          & 1 & 2  & 18.033 & \textbf{0.136} & 3 & 10 & 41.539 \\
    ICOT          & 0.069          & 2 & 3  & 1.885  & 0.079          & 2 & 3  & 9.115  \\
    PCT           & 0.081          & 1 & 2  & 0.004  & 0.081          & 1 & 2  & 0.004  \\
    2-Step        & 0.071          & 3 & 8  & 0.671  & 0.079          & 3 & 8  & 0.507  \\
    \bottomrule
    \end{tabular}
    \end{subtable}
    
\end{table}

\subsection{Results on Clustering Performance}\label{subsec:results}

Table~\ref{tab:ari} presents the results of the following evaluation metrics for the five clustering tree methods: 
\begin{itemize}
    \item \textbf{ARI}: Adjusted Rand index~\cite{Hubert1985}, which is an external evaluation metric for measuring the similarity between constructed clusters and ground-truth clusters;
    \vspace{1mm}
    \item \textbf{Dep}: Depth of a clustering tree;
    \vspace{1mm}
    \item \textbf{\#Cls}: Number of clusters;
    \vspace{1mm}
    \item \textbf{Time (s)}: total computation time in seconds required for model construction including preprocessing;
\end{itemize}
where the best ARI values among the five methods are shown in bold.
The number of clusters was tuned from 2 to 10 using the silhouette coefficient~\cite{Rousseeuw1987} or the Dunn index~\cite{Dunn1974} as internal evaluation metrics. 
Note that the silhouette coefficient assesses the average goodness of fit for each data point, while the Dunn index focuses on the minimum distance between clusters and the maximum diameter within a cluster.

The ICOMT-K method achieved higher ARI values than baseline (i.e., ICOT, PCT, and 2-Step) methods across all datasets.
These results support the superior clustering accuracy of the ICOMT-K method compared to the baseline methods.
On the other hand, the ICOMT-C method achieved the highest ARI values on the Estate dataset but showed low ARI values on the Seeds and Customer datasets.
This suggests that the clustering accuracy was stabilized by our 1D-$K$-means binning method, which can discretize numerical features while considering the data distribution. 

The ICOMT-K method maintained high ARI values without compromising the simplicity of the tree structure.
For example, when using the Dunn index on the Seeds dataset, the ICOMT-K method achieved the highest ARI value with a simple tree of depth 1 and 3 clusters, while the ICOT and 2-Step methods generated trees of depth 3 and 5 clusters.
Furthermore, when using the silhouette coefficient on the Statlog dataset, the ICOMT-K method achieved the highest ARI value with a simple tree of depth 1 and 3 clusters, while the 2-Step method generated a complex tree of depth 3 and 8 clusters.

Compared to other baseline methods, our ICOMT methods required long computation times, particularly on the relatively large Statlog dataset.
This is likely due to the complexity of our ICOMT computational procedure compared to other baseline methods.
However, outside of the Statlog dataset, our ICOMT methods consistently completed computations in less than one second.
Even with the Statlog dataset, their computation time was at most a few minutes.
Considering that the cluster analysis is typically performed offline, this computational cost can be considered acceptable.

Compared to tuning the number of clusters based on the silhouette coefficient, tuning based on the Dunn index tended to increase the number of clusters in some datasets.
For example, in the Statlog dataset, the ICOMT-C method generated 2 clusters when tuned using the silhouette coefficient, but 10 clusters when tuned using the Dunn index.
These results suggest that tuning with the Dunn index, which depends on the extrema of the clusters, tends to select a large number of finer clusters.

\begin{figure}[!t] % 量が多いので [p] (独立ページ) または [!t] を推奨
  \centering

  % --- 1行目 (2個) ---
  \begin{subfigure}{0.59\linewidth}
    \centering
    \includegraphics[width=\linewidth]{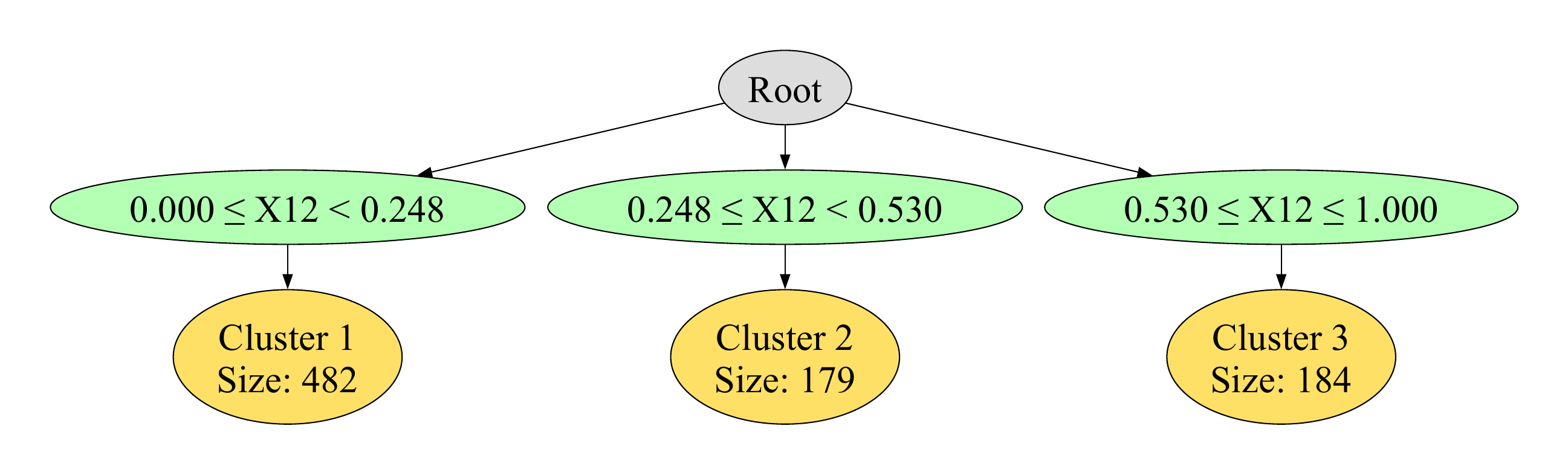}
    \caption{ICOMT-K}
    \label{fig:icomtk_tree}
  \end{subfigure}
  \hfill
  \begin{subfigure}{0.39\linewidth}
    \centering
    \includegraphics[width=\linewidth]{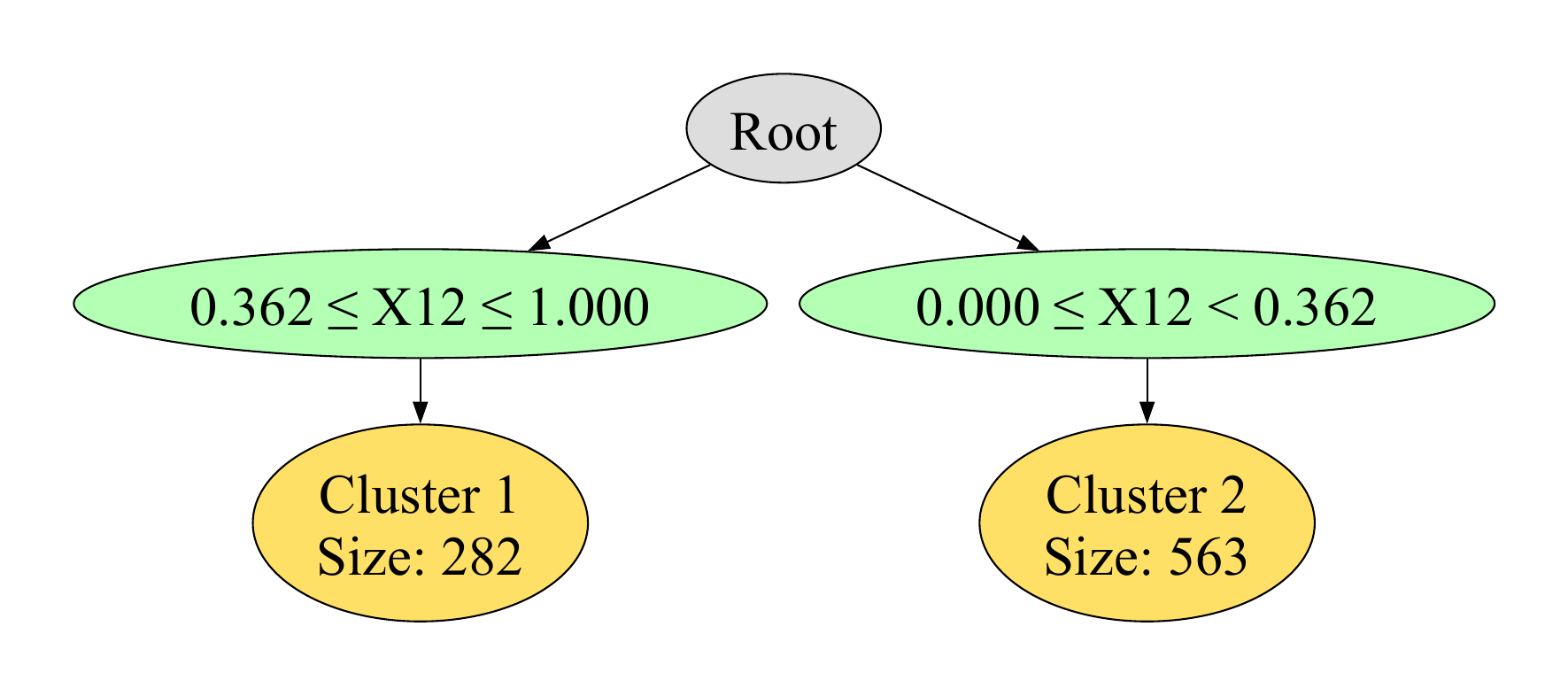}
    \caption{ICOMT-C}
    \label{fig:icomtb_tree}
  \end{subfigure}

  \vspace{10pt} % 行間のスペース調整（空行を入れることで改行されます）

  % --- 2行目 (2個) ---
  \begin{subfigure}{0.49\linewidth}
    \centering
    \includegraphics[width=\linewidth]{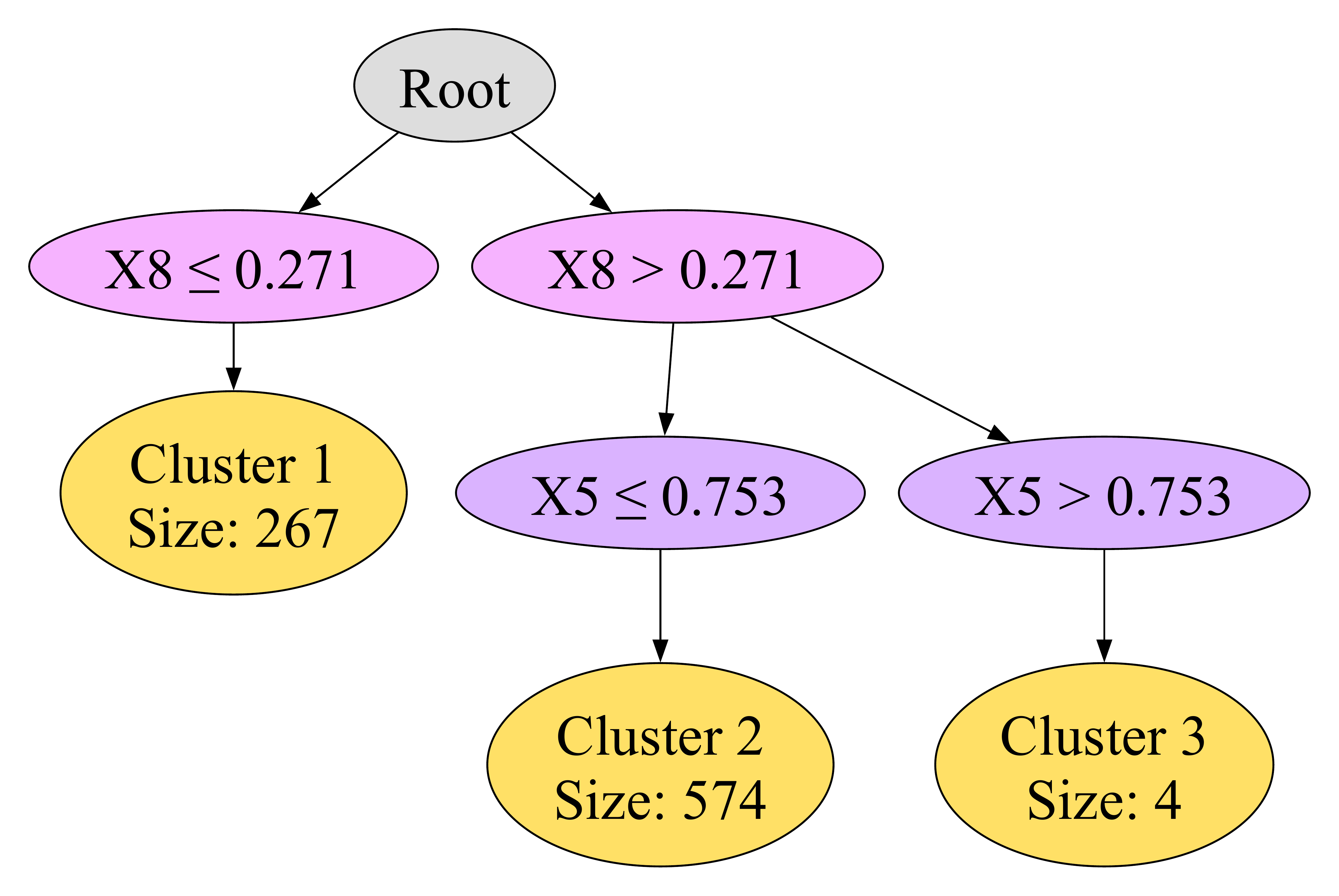}
    \caption{ICOT}
    \label{fig:icot_tree}
  \end{subfigure}
  \hfill
  \begin{subfigure}{0.39\linewidth}
    \centering
    \includegraphics[width=\linewidth]{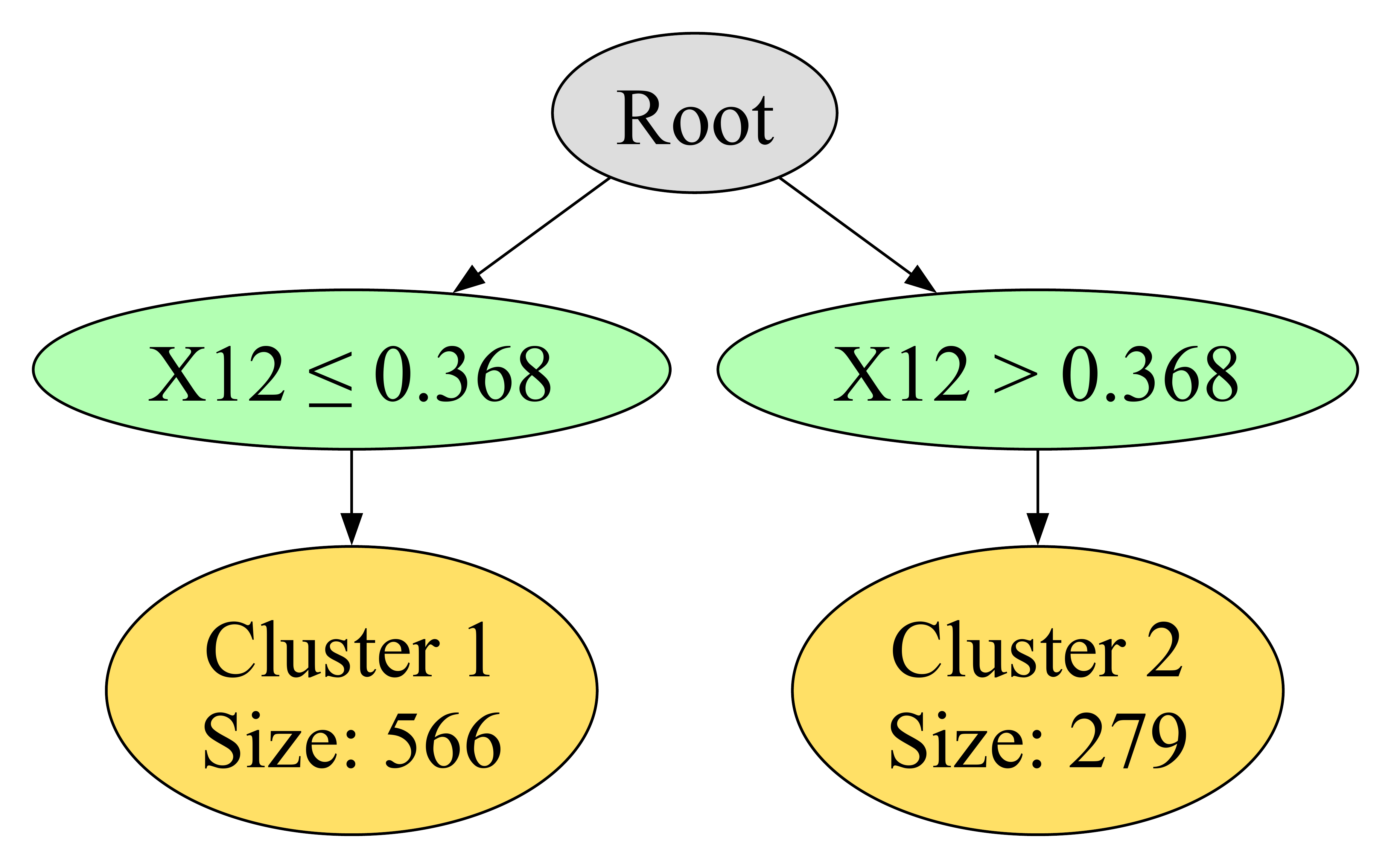}
    \caption{PCT}
    \label{fig:pct_tree}
  \end{subfigure}

  \vspace{10pt}

  % --- 3行目 (1個) ---
  \begin{subfigure}{\linewidth}
    \centering
    \includegraphics[width=0.99\linewidth]{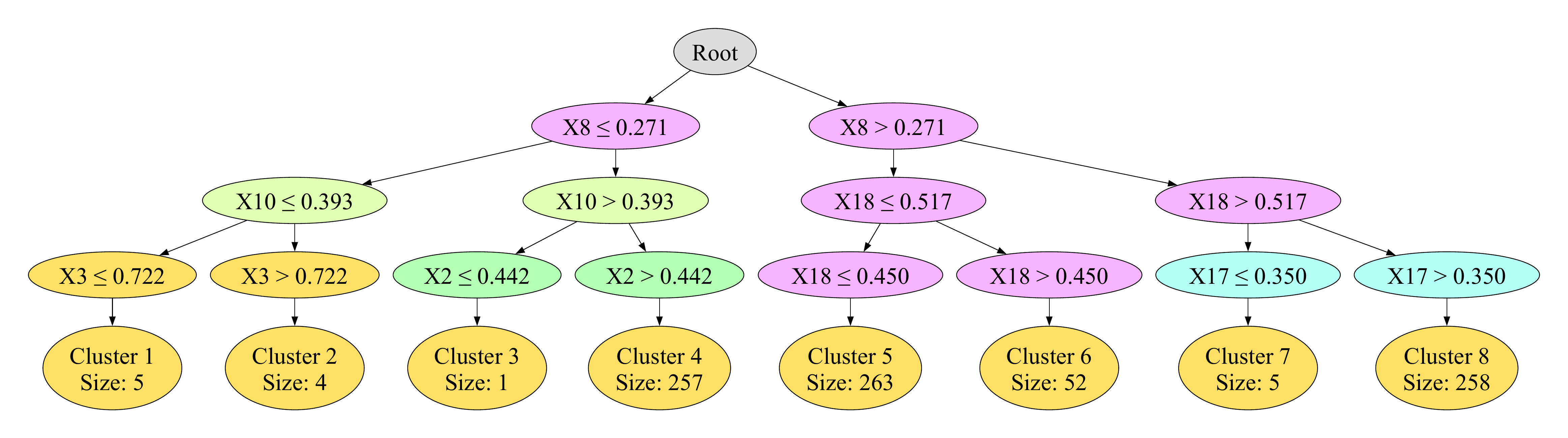}
    \caption{2-Step}
    \label{fig:carthybrid_tree}
  \end{subfigure}

  \caption{Comparison of constructed clustering trees}
  \label{fig:all_trees}
\end{figure}

\subsection{Comparison of Constructed Clustering Trees}\label{subsec:comparison}

Fig.~\ref{fig:all_trees} illustrates clustering trees constructed on the Statlog dataset (cf.~Table~\ref{tab:statlog_features}) using the five methods, with the number of clusters tuned by the silhouette coefficient.

The ICOMT-K and ICOMT-C methods created trees of depth 1, and the PCT method created clusters nearly identical to those of the ICOMT-C method.
In particular, the ICOMT-K method, which employs the multi-way tree structure based on the 1D-$K$-means binning method, achieved high clustering accuracy (ARI) while maintaining a small tree depth (Table~\ref{tab:ari4}).
Additionally, these three methods used the same feature X12 (Scaled variance along the minor axis), suggesting that this feature is likely the primary factor in data separation.

In contrast, the ICOT method created a tree of depth 2 and 3 clusters, and the 2-Step method created a complex tree of depth 3 and 8 clusters.
Since these methods perform sequential splitting based on the binary tree structure, the resultant trees tend to be deep, making it difficult to grasp the overall structure.

\section{Conclusion}\label{sec:conclusion}

We proposed ICOMT, a high-performance computational framework for interpretable clustering based on multi-way decision trees.
Specifically, we applied one-dimensional $K$-means clustering to the discretization of numerical features.
Moreover, by extending the OMT framework~\cite{subramanian2023} designed for supervised learning, we formulated the BLO model for constructing optimal multi-way decision trees for unsupervised clustering tasks. 

We evaluated the effectiveness of our interpretable clustering framework through the experiments using the four publicly available datasets. 
Our method surpassed existing methods (ICOT~\cite{bertsimas2021}, PCT~\cite{Blockeel1998}, and the two-step (clustering and tree construction) scheme~\cite{moshkovitz2020}) in both clustering accuracy and interpretability.
In particular, our 1D-$K$-means binning method frequently showed higher clustering accuracy than the previous cumulative binning method.
Furthermore, the decision trees obtained with our method were shallow and clearly showed the splitting conditions for each cluster, allowing for an intuitive understanding of how each feature contributes to cluster formation.

Our method assumes an exclusive partitioning, where each data point belongs to only one cluster; however, clusters often partially overlap in real-world datasets.
Future research directions include extending our method to overlapping cluster analysis, which allows a single data point to belong to multiple clusters.
Further research directions include exploring more effective methods for discretizing numerical features and designing scalable algorithms for our BLO model for interpretable clustering.
%Developing a procedure to automatically select an appropriate discretization method for each feature is also promising.

% \newpage
% \clearpage
\bibliographystyle{splncs04}
\bibliography{cite.bib} % BibTeX を使う場合

\end{document}